\pdfoutput=1
\documentclass[11pt]{article}

\usepackage[preprint]{acl}

\usepackage{times}
\usepackage{latexsym}

\usepackage[T1]{fontenc}
\usepackage{booktabs}
\usepackage{graphicx}
\usepackage{xcolor,colortbl}

\usepackage[utf8]{inputenc}
\usepackage{colortbl}
\usepackage{xcolor}
\usepackage[most]{tcolorbox}
\usepackage{microtype}

\usepackage{inconsolata}

\usepackage{graphicx}

\title{STRICT: Stress-Test of Rendering Image Containing Text}

\author {
    Tianyu Zhang\textsuperscript{1}\thanks{These authors contributed equally},
    {\bf Xinyu Wang\textsuperscript{2}\footnotemark[1]},
    {\bf Lu Li\textsuperscript{\rm 3}\footnotemark[1]},
    {\bf Zhenghan Tai\textsuperscript{\rm 4}}, \\
    {\bf Jijun Chi\textsuperscript{\rm 4}},
    {\bf Jingrui Tian\textsuperscript{\rm 5}}, 
    {\bf Hailin He\textsuperscript{\rm 6}}
    {\bf Suyuchen Wang\textsuperscript{\rm 1}} \\
    \textsuperscript{\rm 1}Mila, University of Montreal 
    \textsuperscript{\rm 2}McGill University
    \textsuperscript{\rm 3}University of Pennsylvania \\
    \textsuperscript{\rm 4}University of Toronto
    \textsuperscript{\rm 5}University of California, Los Angeles \\
    \textsuperscript{\rm 6}Southwestern University of Finance and Economics}
\begin{document}
\maketitle
\begin{abstract}
While diffusion models have revolutionized text-to-image generation with their ability to synthesize realistic and diverse scenes, they continue to struggle with generating consistent and legible text within images. This shortcoming is commonly attributed to the locality bias inherent in diffusion-based generation, which limits their capacity to model long-range spatial dependencies. In this paper, we introduce \textbf{STRICT}, a benchmark designed to systematically stress-test the ability of diffusion models to render coherent and instruction-aligned text in images. Our benchmark evaluates models across multiple dimensions: (1) the maximum length of readable text that can be generated; (2) the correctness and legibility of the generated text, and (3) the ratio of not following instructions for generating text. We assess several state-of-the-art models, including proprietary and open-source variants, and reveal persistent limitations in long-range consistency and instruction-following capabilities. Our findings provide insights into architectural bottlenecks and motivate future research directions in multimodal generative modeling. We release all our evaluation pipeline at \url{https://github.com/tianyu-z/STRICT-Bench/}.

\end{abstract}

\section{Introduction}
Text-to-image generation has made remarkable strides with the advent of diffusion models ~\cite{DBLP:conf/nips/HoJA20,DBLP:conf/cvpr/RombachBLEO22,DBLP:journals/corr/abs-2211-01324,DBLP:conf/cvpr/FengZYFLCL0YFSC23,DBLP:conf/iclr/GalAAPBCC23,zhang2023texttoimage}, which can now produce highly realistic images from natural language prompts. However, the generation of accurate and coherent text within images, such as complex road signs, product labels, or blackboards, remains a major unsolved problem ~\cite{liu2023characterawaremodelsimprovevisual, chen2023textdiffuserdiffusionmodelstext}. Unlike general object generation, rendering text demands strict spatial precision, character-level continuity~\cite{fallah2025textinvision}, and strong adherence to instruction semantics. Due to their iterative and local sampling nature, diffusion models often fail to maintain global coherence ~\cite{zhang2024artist}, leading to text that is jumbled, misspelled, or visually fragmented. These failures highlight a fundamental challenge in aligning image synthesis with structured linguistic content~\cite{chen2025comprehensive}.

Recent advances, such as OpenAI’s Image-4o~\cite{openai2025gpt4oimage}, have shown promising progress in this domain, achieving near-human performance in rendering embedded text. Similarly, open-source models like HiDream-L1~\cite{hidreami1} and SeedDream 3~\cite{gao2025seedream} have reported comparable success in overcoming long-range dependency issues. Yet, a systematic and quantitative evaluation of these capabilities remains lacking.

In this work, we present \textbf{STRICT} (Stress-Test of Rendering Image Containing Text), a comprehensive benchmark designed to rigorously evaluate the performance of diffusion models in generating image-embedded text. Our contributions are threefold:
\begin{itemize}
    \item We introduce a multi-lingual benchmark that tests model performance on rendering texts of varying lengths in English, Chinese, and French.
    \item We propose quantitative metrics for assessing (1) the maximum readable text length; (2) the correctness of the generated content, and (3) the ratio of not following instructions for generating text.
    \item We analyze recurring failure modes, including truncation in longer texts and the inability to follow explicit textual instructions.
\end{itemize}

Through this evaluation, we aim to expose current limitations, identify failure patterns, and guide the development of structure-aware generative models capable of producing semantically and visually coherent text within images.

% MAIN Figure Here

\section{Task Design}
\label{sec:task_design}

\begin{figure*}[t]
    \centering
    \includegraphics[width=1\linewidth]{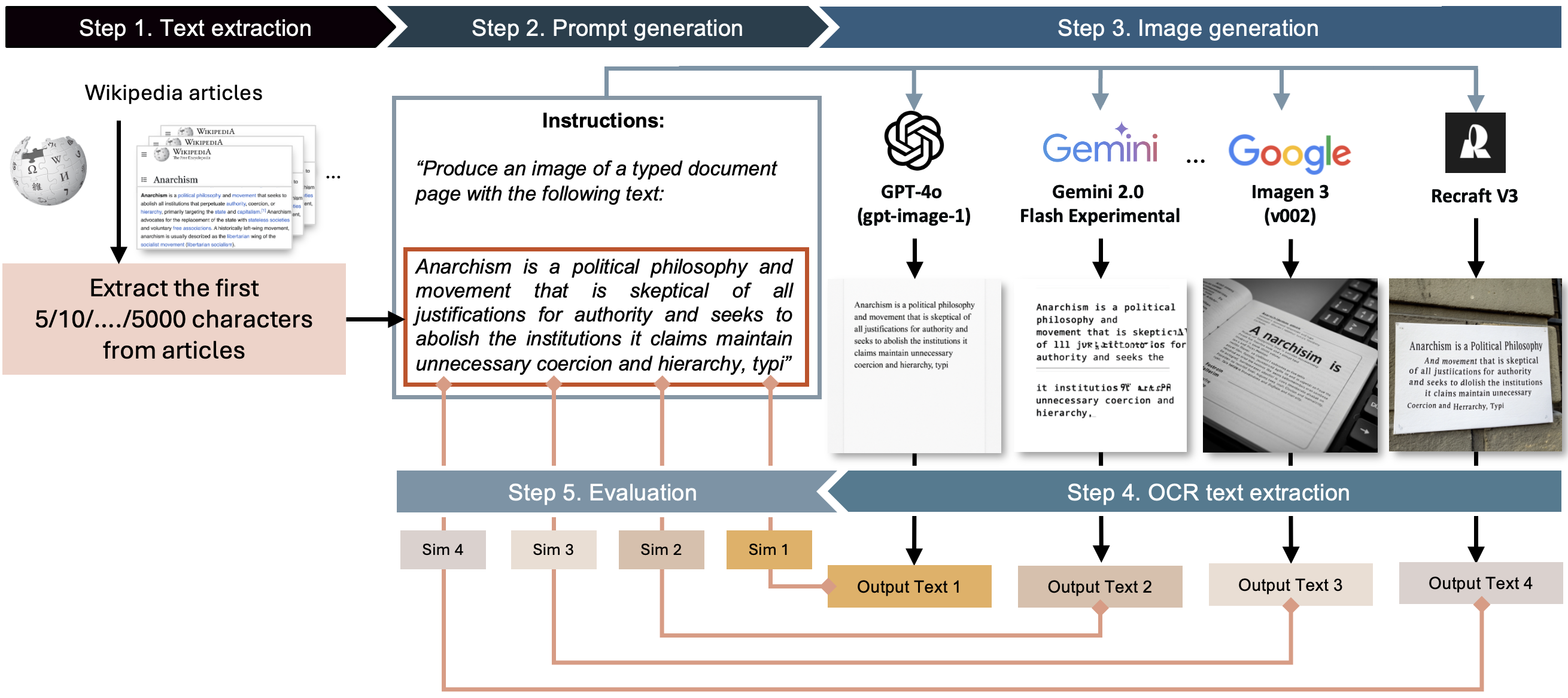}
    \caption{Illustration of the dataset creation and evaluation pipeline for \textbf{STRICT}. \textbf{Step 1:} We begin by selecting multilingual text samples from Wikipedia~\cite{wikidump}, and extracting character sequences of varying lengths, ranging from 5 to 5000 characters.
\textbf{Step 2:} For each text sample, we generate natural language instructions prompting models to create ``a plain Word document with black text on a white background, without decorative elements,'' embedding the extracted text from Step 1.
\textbf{Step 3:} These prompts are then passed to various text-to-image generation models to produce the corresponding output images.
\textbf{Step 4:} Optical character recognition (OCR) is applied to the generated images to extract the rendered text.
\textbf{Step 5:} Finally, we evaluate the quality of the rendered text by comparing the OCR output to the ground truth using similarity metrics, including normalized edit distance (NED), character error rate (CER), and word error rate (WER).  
    }
    \label{fig:main}
\end{figure*}

The primary objective of our benchmark is to rigorously evaluate the capability of text-to-image generation models to render accurate and coherent text embedded within synthetic images. The task is to utilize the model to be evaluated with a ground truth content string and instructed with a natural language prompt template: \textit{f"Produce an image of a typed document page with the following text: } \textbf{[TEXT]} \textit{"} to generate an image containing this specific text \textbf{[TEXT]}. The characteristics of the ground truth text sample are systematically varied to probe different aspects of model performance. These variations include:
\begin{itemize}
    \item \textbf{Text Length:} Ground truth texts range from short phrases to longer paragraphs to determine the maximum length of text a model can generate coherently. This helps assess how accuracy degrades as the quantity of text increases.
    \item \textbf{Language:} Texts are selected from Wikipedia~\cite{wikidump} based on different languages. 
\end{itemize}

For each sample, a model generates an image based on the input prompt containing the target ground truth text and then forms a pair with its corresponding ground truth text file. We collect these generated image-text pairs for evaluating each model.

\section{Evaluation Metrics}
\label{sec:evaluation_metrics}

To quantify the performance of models in rendering text, we employ an OCR-based verification framework. The text content from each generated image is first extracted using an OCR engine, followed by a comparison against the original ground truth text.

\subsection{OCR and Preprocessing}
We utilize the Tesseract OCR engine~\cite{tesseract} for extracting text from generated images. Tesseract is a widely adopted open-source OCR tool capable of recognizing over 100 languages and offering configurable page segmentation modes (PSM) to suit different layout scenarios. In our evaluation, we primarily use English, French, and Chinese language models with PSM set to 3, which indicates fully automatic page layout analysis.

To ensure fair and robust evaluation, we perform minimal but critical preprocessing of both the OCR-extracted text and the ground truth references. Specifically, we adopt a strict text processing strategy that normalizes whitespace across both texts, collapsing all forms of whitespace (spaces, tabs, newlines) into a single space, and trims leading and trailing whitespace.

After preprocessing, we evaluate OCR performance using a suite of metrics: character-level accuracy (Character Error Rate, CER) ~\citep{radford2023robust, conneau2021unsupervised}, word-level accuracy (Word Error Rate, WER) ~\citep{morris2004interspeech, kim2021semantic}, normalized edit distance (NED) ~\citet{fisman2022normalized}, and sequence similarity using Python package \texttt{difflib} ~\cite{difflib}. For a more granular analysis, we compute these metrics under two settings: a full comparison, which compares the entire OCR output to the full ground truth, and a truncated comparison, which compares the OCR output against a truncated version of the ground truth matched to the number of tokens recognized by OCR. This dual-mode evaluation allows us to assess not only absolute performance but also the model’s ability to preserve textual order and correctness under realistic generation length constraints.

Through these evaluations, we reveal systematic differences in how text-to-image models perform under varying linguistic and spatial constraints in multi-language settings.

\subsection{Quantitative Metrics}
Following OCR extraction and preprocessing, the recognized text is compared against the ground truth using several order-preserving metrics. These metrics assess the accuracy of the generated text at both word and character levels. We report "Full" and "Truncated" versions for each metric; the "Full" version compares against the entire ground truth, while the "Truncated" version may adapt the comparison based on the length of the shorter sequence (typically the OCR output), providing insight into partial correctness. 

\begin{enumerate}
    % \item \textbf{Sequence Similarity (SqSim):} This metric measures the ratio of matching sequences to the total number of elements in both sequences. It ranges from 0.0 (no match) to 1.0 (perfect match), with higher values indicating better performance.
    \item \textbf{Normalized Edit Distance (NED):}
    \label{NED}
    NED quantifies the dissimilarity between the ground truth and OCR output based on character-level Levenshtein distance (edit distance). We adopt the Levenshtein distance normalization proposed in~\citet{fisman2022normalized}, where the cost of inserts, deletes and swaps are all 1, and the normalization factor is the length of the optimal edit path. This process results in a score between 0.0 (identical strings) and 1.0. Lower NED values indicate greater similarity. In our observation, some models consistently generate words with typos, making character-level metrics better reflect the actual generation performance.

    \item \textbf{Character Error Rate (CER):} Other than NED, we also use two commonly used metrics in speech recognition: Character Error Rate (CER)~\citep{radford2023robust, conneau2021unsupervised} and Word Error Rate (WER)~\citep{morris2004interspeech, kim2021semantic}. Similar to NED, CER operates at the character level. It is more sensitive to minor errors like single incorrect letters or OCR misrecognitions. CER is also normalized by the number of characters in the ground truth (for Full CER), with lower values being better. Note that the CER values do not have an upper limit.
    % CER = (S + D + I) / M, where M is number of characters in reference.

    \item \textbf{Word Error Rate (WER):} WER is a standard word-level metric commonly seen in speech recognition and OCR, calculating the minimum number of word-level insertions, deletions, and substitutions required to transform the OCR output into the ground truth text. The result is typically normalized by the number of words in the ground truth text (for Full WER). Lower WER values signify higher accuracy. Note that the WER values do not have an upper limit.
    % WER = (S + D + I) / N, where N is number of words in reference.

    \item \textbf{Ratio of Not Following Instructions (RNFI):} RNFI measures the extent to which a model fails to follow the prompt by generating discretionary natural images instead of faithfully rendering the given text. For each sample, we compute the ratio between the number of characters extracted from the model-generated image (via OCR) and the number of characters in the ground-truth input text. We then count the number of samples where this ratio falls below 1\%, indicating that the model has effectively ignored the instruction, i.e., less than 1\% of the characters from the prompt are present in the generated image, \textbf{regardless of their correctness}.
\end{enumerate}

For each metric, we calculate aggregate statistics, including mean, bootstrapped standard deviation for mean, and bootstrapped confidence intervals, across the entire dataset of evaluated image-text pairs. This provides a robust overall assessment of a model's text generation capabilities under the specified evaluation conditions. The length of the ground truth text and the OCR-extracted text in characters are also recorded to provide context for the error rates.

\section{Experiment Results}
\label{sec:experiment_results}

We evaluate a diverse set of state-of-the-art text-to-image generation models, including proprietary and open-source variants, on the \textbf{STRICT} benchmark. The models tested include: GPT-4o (\texttt{gpt-image-1}) ~\cite{openai2025gpt4oimage}, Seedream 3.0~\cite{gao2025seedream}, Recraft V3~\cite{recraftv3}, HiDream-I1-Dev~\cite{hidreami1}, Imagen 3 (\texttt{imagen-3.0-generate-002})~\cite{imagen3}, FLUX 1.1 pro~\cite{flux1.1}, Gemini 2.0 (\texttt{gemini-2.0-flash-preview-image-generation}) ~\cite{gemini2.0}, as well as open-source models such as Stable Diffusion 3.5 Medium~\cite{stablediffusion3.5}, Anytext 2~\cite{DBLP:journals/corr/abs-2411-15245}, and the \texttt{TextDiffuser 2}~\cite{DBLP:conf/nips/ChenHLCWC23,DBLP:conf/eccv/ChenHLCWC24}. We also include partial results for \texttt{Qwen-Image}~\cite{wu2025qwenimagetechnicalreport} and \texttt{gemini-2.5-flash (nano-banana)}~\cite{Carter2025_master_nano_banana_prompt}, which is released after we submit the paper. The text dataset is sourced from Wikipedia~\cite{wikidump}, which is licensed under the Creative Commons Attribution-ShareAlike 3.0 (CC-BY-SA 3.0) license. The OCR processing was performed using Tesseract OCR~\cite{tesseract}, which is distributed under the Apache License 2.0. Both licenses permit use for research and commercial purposes, provided the respective attribution and license terms are followed.

\begin{figure*}[t]
    \centering
    \includegraphics[width=1\linewidth]{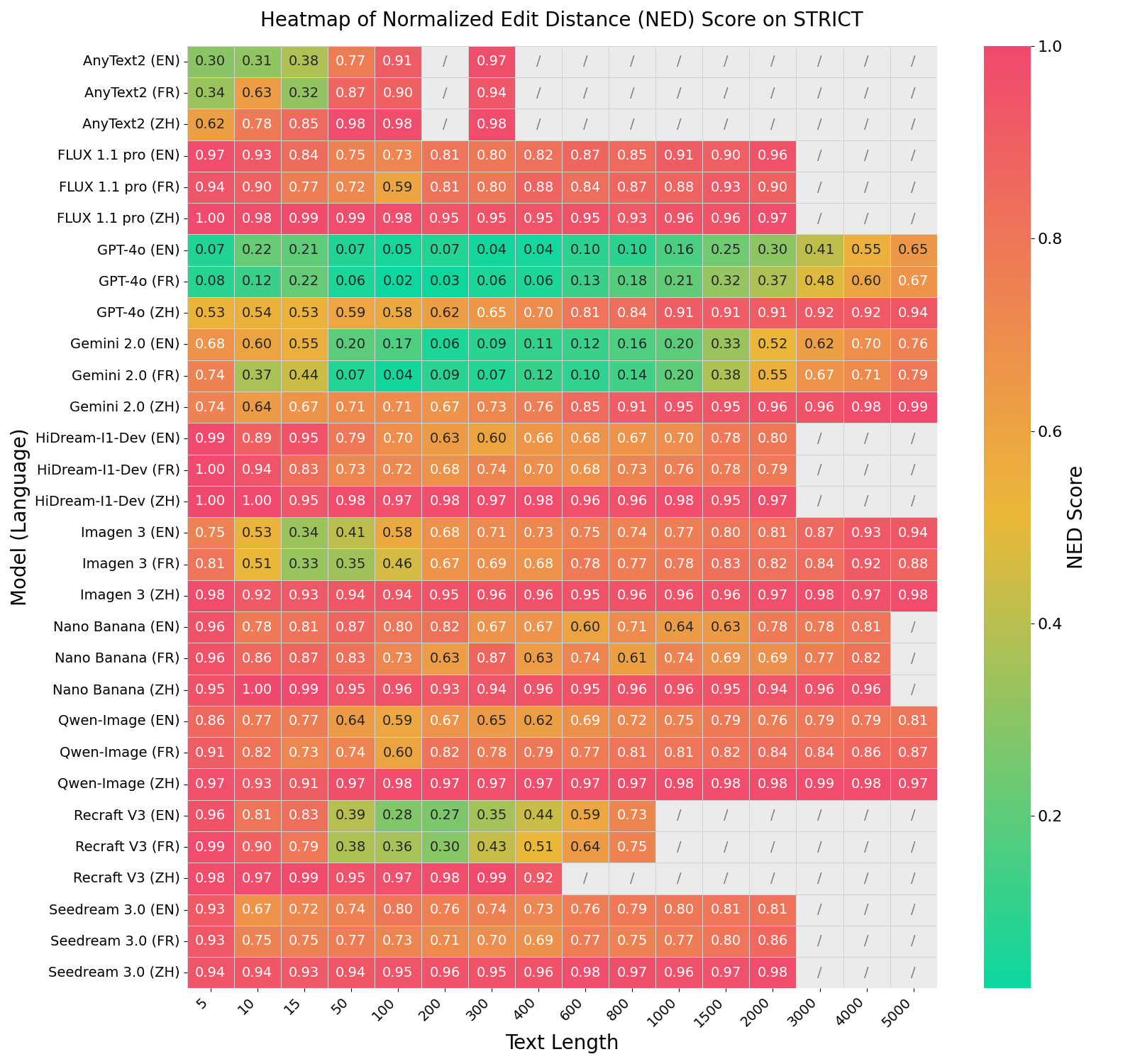}
    \caption{Heatmap of Normalized Edit Distance (NED) scores on the \textbf{STRICT} benchmark. Models are evaluated across three languages (EN: English, FR: French, ZH: Chinese) and varying text lengths. Lower NED scores (green) indicate better performance (higher text rendering accuracy). Grey cells denote untested lengths, as models were evaluated on ranges corresponding to their capabilities or API limitations.}
    \label{fig:heatmap}
\end{figure*}

\begin{figure*}[t]
    \centering
    \includegraphics[width=1\linewidth]{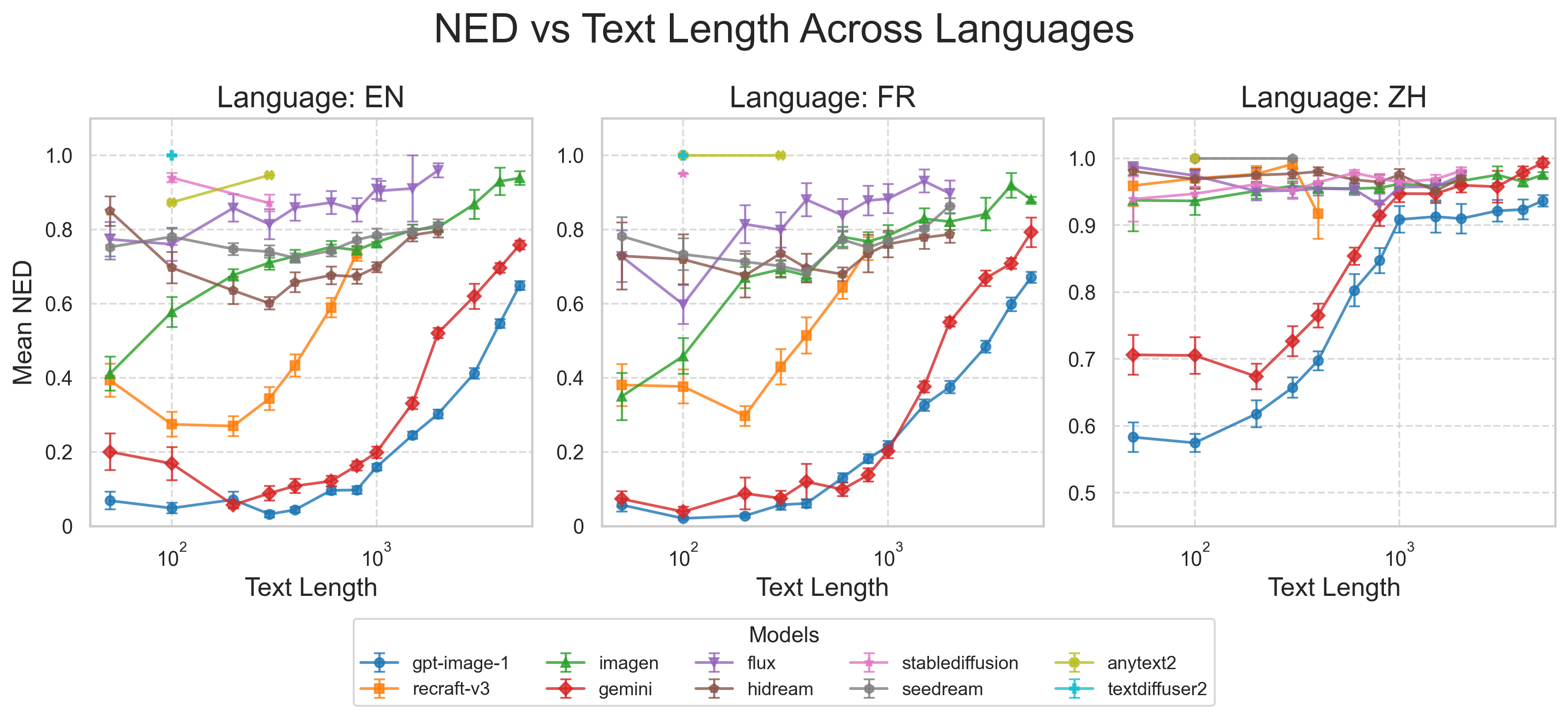}
    \caption{
    \textbf{Normalized Edit Distance (NED) vs. Text Length across Languages.} 
    We evaluate ten state-of-the-art text-to-image generation models on multilingual text rendering using English (EN), French (FR), and Chinese (ZH) excerpts sampled from Wikipedia, with input lengths ranging from 5 to 5000 characters. Each model is prompted with identical semantic content across varying lengths, and OCR is applied to the generated images to compute character-level NED scores. Higher-performing models such as GPT-4o, Gemini 2.0, and Imagen 3 are evaluated up to 5000 characters, while Stable Diffusion 3.5, AnyText2, and TextDiffuser2 are evaluated up to 300 characters, and the remaining models up to 2000. Lower NED scores indicate better text fidelity and layout consistency.
    }
    \label{fig:NED}
\end{figure*}

\paragraph{Overall Performance.} 
According to Figure \ref{fig:NED}, across all evaluated languages and text length, GPT-4o and Gemini-2.0 outperform all competing models significantly by a large margin in terms of character accuracy, word accuracy, and instruction adherence. We evaluated ten models with varying capabilities. For the weakest models (Anytext 2, TextDiffuser 2, Stable Diffusion 3.5 Medium), we tested input lengths ranging from 5 to 300 characters. For moderately performing models (FLUX 1.1 pro, Seedream 3.0, HiDream-I1-Dev), the tested character lengths ranged from 50 to 2,000. Although moderate performance, the API upper limit of Recraft V3 is 1000 bytes (1,000 Latin characters or 500 Chinese characters). Thus, we test Recraft V3 from 50 to its upper limits. Moreover, the strongest models (Imagen 3, Gemini 2.0 and GPT-4o) were evaluated on inputs ranging from 50 to 5,000 characters. For both English and French, GPT-4o and Gemini maintain strong performance up to approximately 800 characters, beyond which accuracy begins to decrease. Detailed CER and WER metrics are displayed in the Appendix \ref{sec:appendix2} for reference. For Chinese, the overall model performance remains poor. However, GPT-4o still consistently outperforms the other models. We display all the scores in the form of heatmap in Figure \ref{fig:heatmap}. For detailed scores with standard deviation, please check Table \ref{tab:ned}.

\begin{figure*}[t]
    \centering
    \includegraphics[width=1\linewidth]{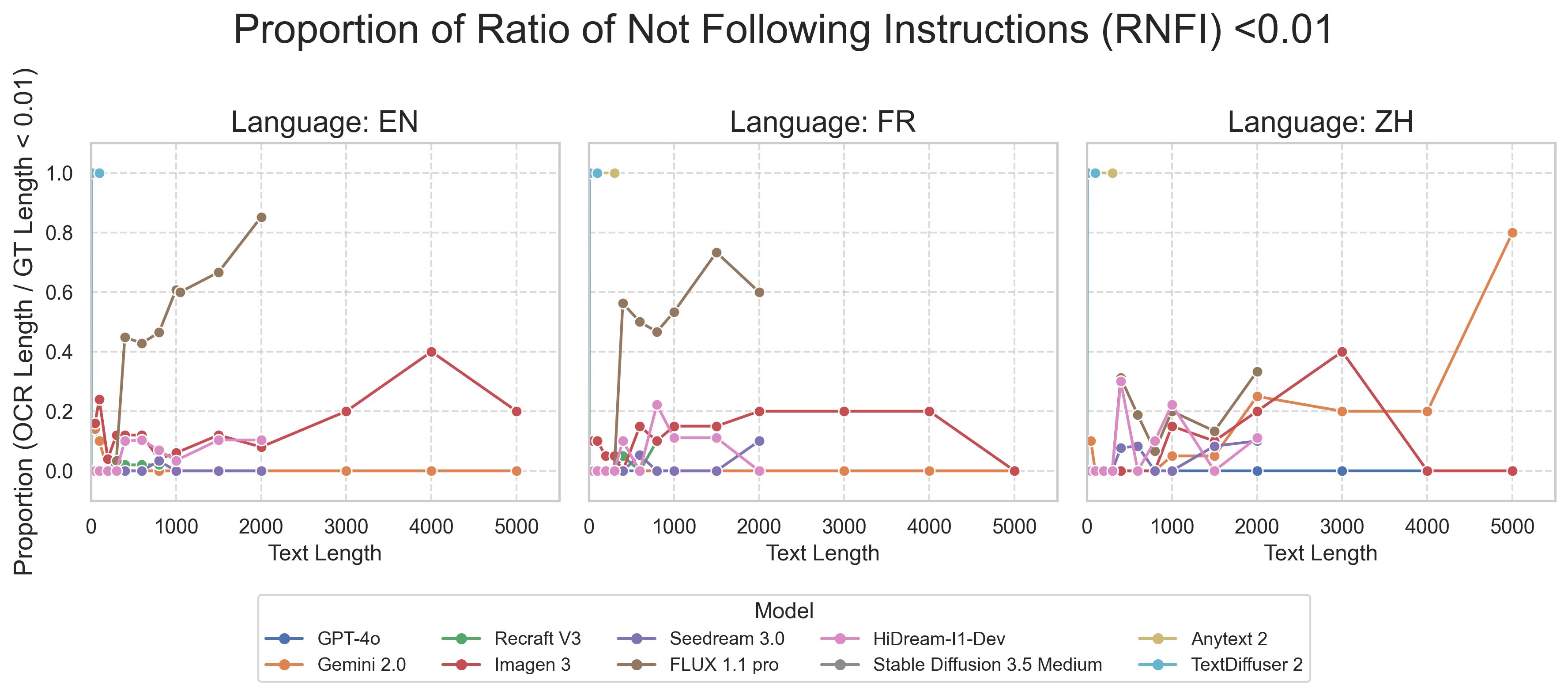}
    \caption{
\    \textbf{Instruction Following Failure Rate across Text Lengths.} We plot the Ratio of Not Following Instructions (RNFI), defined as the percentage of samples where a model fails to render the input text. A failure is recorded if the ratio of characters in the generated image (measured via OCR) to characters in the input text is less than 1\%. This metric captures catastrophic failures, such as generating a natural image instead of text, and does not penalize minor rendering errors. Evaluations use multilingual text (EN, FR, ZH) from Wikipedia with input lengths from 5 to 5,000 characters. Lower values indicate better robustness and instruction adherence.}
    
    \label{fig:rnfi}
\end{figure*}

\paragraph{Ratio of Not Following Instructions.} 
From Figure \ref{fig:rnfi}, we can see that some models including Flux 1.1 pro and Gemini 2.0 tend not to follow the instructions, especially when the number of characters in the given prompt becomes longer. We will discuss more in Section \ref{sec:discussion}.

\section{Discussion}
\label{sec:discussion}

\begin{figure*}[t]
    \centering
    \includegraphics[width=1\linewidth]{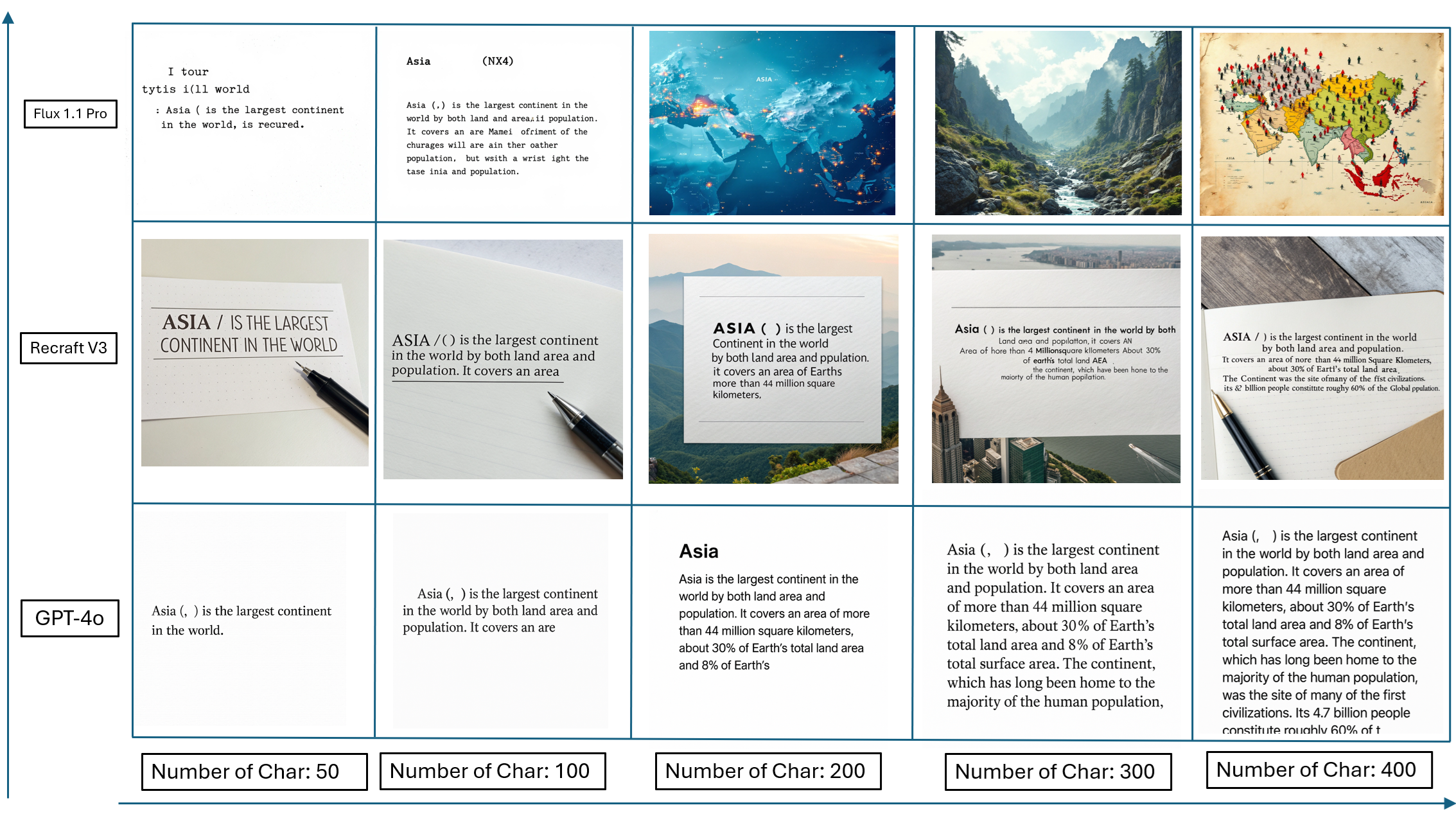}
    \caption{Case study on instruction-following failures. The three rows correspond to Flux 1.1 Pro, Recraft V3, and GPT-4o. GPT-4o consistently adheres to the given instructions, whereas Flux 1.1 Pro increasingly ignores them as the character count grows. Recraft V3 continues to generate text within the image but introduces background elements that were not requested in the prompt.}
    \label{fig:case_study}
\end{figure*}

Our study highlights both the capabilities and limitations of current text-to-image models in faithfully rendering structured textual content. While models like GPT-4o ~\cite{openai2025gpt4oimage} demonstrate impressive gains and establish a new standard, the majority of diffusion models ~\cite{gao2025seedream,recraftv3,hidreami1,imagen3,flux1.1,stablediffusion3.5,DBLP:conf/nips/ChenHLCWC23,DBLP:conf/eccv/ChenHLCWC24} still face significant challenges in instruction-following, text alignment, and multi-lingual generalization.

\paragraph{Performance Degradation with Text Length.} 
Most diffusion models demonstrate a marked decline in performance as the input text length exceeds approximately 200 characters. This threshold aligns with the 77-token limit of the CLIP text encoder, which is used in HiDream, stable diffusion series and other diffusion models. Taking a step back, as mentioned in \cite{zhang2024long0clip0}, the real effective token length of CLIP is 20 rather than 77. Considering the above two points together, it is likely that the model's ability to capture and condition on longer instructions is restricted. Moreover, existing training corpora rarely include such long and instruction-heavy prompts, further exacerbating generalization challenges in these regimes.

\paragraph{Instruction-Following Failures.} 
With longer text prompts, diffusion models increasingly fail to adhere to instruction semantics. Instead of generating a document-like image containing the target text, many models instead synthesize a naturalistic image related to the subject matter of the prompt, completely omitting any embedded text. For example, in Figure \ref{fig:case_study}, when instructed to render a document stating ``Asia ( ,  ) is the largest continent in the world by both land area and population....", some models, especially Flux 1.1 pro, return an illustration of an Asian map instead of the text itself as the number of characters increase. We hypothesize that this failure arises from three main causes: (1) the dilution of the instruction signal in long prompts, which reduces the model’s focus on the ``document should contain the text:" instruction; (2) the limited capacity of CLIP-based or similar text encoders to capture long-range logical structure in extended inputs. These encoders may process such inputs more like a bag-of-words and thus compromise instruction fidelity; (3) the use of certain models, such as HiDream-L1, which incorporate LLMs (e.g., Llama-3.1-8B~\cite{grattafiori2024llama}) as encoders, although LLMs are primarily designed as decoders and are not well-suited for encoding tasks. Prior work~\cite{huang2024llm2clip0, behnamghader2024llm2vec0} has highlighted that LLMs generally underperform in encoder roles, although some modifications can improve their effectiveness.

\paragraph{Cross-Lingual Variation.}
We observe consistent differences in performance across languages. English generally yields the highest accuracy, followed by French, while Chinese exhibits the lowest performance. This is likely attributable to insufficient training data for Chinese, rather than intrinsic difficulties in rendering Chinese characters. Structurally, we do not think there is a fundamental reason showing Chinese characters should be harder to generate than Latin alphabet. The performance gap underscores a need for broader multilingual data inclusion in models' pertaining.

% First, we emphasize the importance of architectural and training data innovations that explicitly encourage long-range dependency modeling. Solutions may include hierarchical instruction encoders, longer context-aware transformer backbones, or hybrid encoder-decoder frameworks that disentangle visual scene planning from text rendering.

% Second, we advocate for broader multilingual training, particularly for underrepresented languages like Chinese. Current models underperform primarily due to data scarcity rather than model inadequacy. Addressing this imbalance will be essential for developing equitable multimodal generation systems.

% Finally, we encourage the adoption of STRICT as a standardized benchmark for evaluating text rendering capabilities in generative models. By isolating text generation as a diagnostic challenge, STRICT enables deeper analysis of failure modes and model behavior under diverse task constraints. We release all evaluation code, prompts, and result dashboards to facilitate reproducibility and encourage further research in this area.

\section{Related Work}

\subsection{Diffusion Models}
Diffusion models have emerged as a dominant paradigm in text-to-image synthesis, surpassing traditional generative models like GANs and VAEs in generating high-fidelity and diverse images. Foundational works such as DDPM~\cite{DBLP:conf/nips/HoJA20} and Latent Diffusion Models~\cite{DBLP:conf/cvpr/RombachBLEO22} laid the groundwork for this progress. Subsequent models like EDIFI~\cite{DBLP:journals/corr/abs-2211-01324}, ERNIE-ViLG 2.0~\cite{DBLP:conf/cvpr/FengZYFLCL0YFSC23}, and Textual Inversion~\cite{DBLP:conf/iclr/GalAAPBCC23} have further enhanced the alignment between generated images and textual prompts.

Recent advancements, Stable Diffusion series ~\cite{DBLP:conf/cvpr/RombachBLEO22,DBLP:conf/icml/EsserKBEMSLLSBP24}, have integrated large language models like T5~\cite{DBLP:journals/jmlr/RaffelSRLNMZLL20} to better encode textual information. Despite these improvements, challenges remain in rendering complex, multi-line, or structured text within images~\cite{DBLP:conf/nips/Zhao00BHYW23}.

\subsection{Autoregressive Models}

Autoregressive models offer an alternative approach to image generation by modeling images as sequences of discrete tokens. Recent developments have focused on enhancing spatial consistency and instruction adherence. VAR~\cite{DBLP:conf/nips/TianJYPW24} employs multi-resolution next-token prediction to improve long-range coherence. InstructCV~\cite{DBLP:conf/iclr/GanPSPA24} frames various visual tasks within a unified text-guided generation framework using multi-modal prompts. Models like SeedX~\cite{DBLP:journals/corr/abs-2404-14396} and Chameleon~\cite{DBLP:journals/corr/abs-2405-09818} unify text and image sequences within the same autoregressive framework, enhancing fluency and cross-modal alignment. Additionally, LlamaGen~\cite{DBLP:conf/cvpr/LuCL0KMHK24} adopts refined tokenization and pretrained language modeling to narrow the performance gap with diffusion-based models.

\subsection{Models on Text Rendering}

Rendering accurate and structured text within images remains a core challenge for generative models. To address spatial precision and layout constraints, GlyphControl~\cite{DBLP:conf/nips/YangZL23} enables user-guided glyph placement, while GlyphDraw~\cite{DBLP:conf/mm/MaZ23} and TextDiffuser~\cite{DBLP:conf/nips/ChenHLCWC23} adopt keyword-driven generation and layout masks for structured rendering. TextDiffuser-2~\cite{DBLP:conf/eccv/ChenHLCWC24} further incorporates layout planning and line-level encoding for improved diversity.

Models such as Recraft and the AnyText series~\cite{DBLP:journals/corr/abs-2311-03054,DBLP:journals/corr/abs-2411-15245} enhance multilingual and stylistic versatility by supporting diverse languages and font styles. Meanwhile, methods such as Glyph-SDXL~\cite{DBLP:journals/corr/abs-2403-09622}, Glyph-SDXL-v2~\cite{DBLP:journals/corr/abs-2406-10208}, and GlyphDraw2~\cite{DBLP:conf/aaai/MaZ24} use OCR-guided glyph representations to improve layout accuracy and visual coherence.

Character-level approaches such as DiffSTE~\cite{DBLP:conf/iclr/ZhangL24}, UDiffText~\cite{DBLP:conf/eccv/ZhaoL24}, and Brush Your Text~\cite{DBLP:conf/acl/ZhangL24} refine alignment through attention-based interventions.

Recent models have further advanced text rendering capabilities. Recraft V3 demonstrates proficiency in generating images with long texts and diverse styles~\cite{recraftv3}. HiDream-I1-Dev, an open-source model with 17B parameters, achieves high-quality image generation with prompt adherence~\cite{hidreami1}. Imagen 3, Google's latest model, offers improved detail and text rendering~\cite{imagen3}. FLUX 1.1 pro delivers enhanced composition and artistic fidelity~\cite{flux1.1}. Gemini 2.0 integrates multimodal inputs for native image generation~\cite{gemini2.0}. Stable Diffusion 3.5 introduces a Multimodal Diffusion Transformer architecture, improving typography and complex prompt understanding~\cite{stablediffusion3.5}. SeedDream 3.0~\cite{gao2025seedream}, a strong open-source diffusion model, demonstrates competitive accuracy and layout consistency in rendering multi-line and structured text.

\subsection{Text-to-Image Benchmarks}

Recent work has introduced specialized benchmarks to systematically evaluate T2I models' abilities to render readable, instruction-aligned, and multilingual text. TIFA~\cite{hu2023tifa} assesses semantic faithfulness via QA-based probing, while TypeScore~\cite{sampaio2024typescore} evaluates OCR-based text fidelity and instruction following.

TextInVision~\cite{fallah2025textinvision} addresses structural challenges by varying prompt lengths and complexities to assess how diffusion models handle diverse textual inputs. MARIO-Eval~\cite{chen2023textdiffuser}, built upon the extensive MARIO-10M dataset, offers a large-scale OCR benchmark for evaluating text rendering quality. For multilingual evaluation, AnyText~\cite{tuo2024anytext} introduces a dataset and metrics that encompass various languages and font styles.

Comprehensive benchmark suites like HEIM~\cite{lee2023holistic} and HRS-Bench~\cite{bakr2023hrs} jointly assess image quality, text rendering, and prompt adherence. LenCom-Eval~\cite{lakhanpal2025refining} focuses on long-form prompts to expose generation degradation.

Recent architectural evaluations show LLM-grounded generation~\cite{lian2024llm} significantly improves prompt alignment and cross-lingual generalization. TextMatch~\cite{luo2024textmatch} refines outputs through multimodal feedback from VQA and LLMs. Collectively, these efforts lay a solid foundation for benchmarking advanced T2I systems under realistic, structured, and semantically rich prompts.

\section{Conclusion}
We introduced \textbf{STRICT}, a comprehensive benchmark for evaluating the ability of text-to-image models to render accurate, instruction-aligned, and multilingual text. Our evaluation reveals that while recent models like GPT-4o and Gemini 2.0 show strong performance, most open-source diffusion models still struggle with long-range coherence and instruction fidelity. We highlight key failure modes such as instruction neglect and language-specific disparities. By providing standardized tasks and metrics, \textbf{STRICT} enables targeted diagnosis and guides future improvements in multimodal generation systems.

\section*{Limitations}
Firstly, despite nearly two decades of continuous development, the Tesseract OCR engine~\cite{tesseract} still encounters failure cases in which humans can easily recognize the text. These limitations remain challenging until we can fully overcome the drawbacks of OCR technologies. 

Furthermore, if \textbf{STRICT} becomes a widely adopted benchmark, there is a risk that future models may be fine-tuned or hard-coded to perform well specifically on the dataset’s structure and instructions. This undermines the benchmark's utility as an unbiased generalization test and could lead to inflated leaderboard results without corresponding real-world gains.

\newpage
% \section*{Acknowledgments}
\bibliography{custom}
\newpage
\appendix
\section{Appendix: Prompt Variants}

To test the robustness of our benchmark against prompt phrasing, we experimented with a set of diverse but semantically equivalent instructions for generating document-style images. These prompt variants yielded consistent performance trends across multiple models. To standardize our evaluation pipeline, we selected a single representative prompt (highlighted in red below) for all metric-based experiments:

\begin{tcolorbox}[title=Prompt Variants Explored, rounded corners, colframe=gray!50, colback=gray!5, boxrule=0.7pt, boxsep=4pt, enhanced, toprule=0.7pt]
\begin{itemize}

    \item \textit{Generate a scanned document image with following text: \textbf{[TEXT]}}
    \item \textit{Create a mockup of a scanned document containing the text: \textbf{[TEXT]}}
    \item \textit{Design a sample document scan with the following text: \textbf{[TEXT]}}
    \item \textit{Generate an image of printed note that include these text: \textbf{[TEXT]}}
    \item \textcolor{red}{\textit{Produce an image of a typed document page with the following text: \textbf{[TEXT]}}} % ← selected
    \item \textit{Generate a document scan visualization showing this text: \textbf{[TEXT]}}
    \item \textit{Produce a sample of how a scanned memo might look with this text: \textbf{[TEXT]}}
    \item \textit{Generate an image of a plain Word document with black text on white background without decorative elements, document should contain the text: \textbf{[TEXT]}}
\end{itemize}
\end{tcolorbox}

We observed no significant variation in performance across these prompts, reinforcing the robustness of our task design. For all reported experiments, we standardized on the red-highlighted prompt.
\section{Appendix: Detailed NED Scores in table}
We present the detailed NED scores and corresponding standard deviation in table \ref{tab:ned}.

\begin{table*}[]
\renewcommand{\arraystretch}{1.3} 
\resizebox{\textwidth}{!}{%
\begin{tabular}{@{}lcccccccccccc@{}}
\toprule
\begin{tabular}[c]{@{}c@{}}\textbf{Text}\\ \textbf{Length}\end{tabular} & \textbf{GPT-4o} & \textbf{Gemini 2.0} & \textbf{Recraft V3} & \textbf{Imagen 3} & \textbf{Seedream 3.0} & \textbf{FLUX 1.1 pro} & \textbf{HiDream-I1-Dev} & \begin{tabular}[c]{@{}c@{}}\textbf{Stable Diffusion}\\ \textbf{3.5 Medium}\end{tabular} & \textbf{Anytext 2} &  \textbf{TextDiffuser 2} &  \textbf{Qwen-Image} & \textbf{nano-banana}\\ \midrule 
\multicolumn{13}{c}{EN} \\  \midrule
5 & $0.07 \pm 0.03$ & $0.68 \pm 0.06$ & $0.96 \pm 0.01$ & $0.75 \pm 0.05$ & $0.93 \pm 0.03$ & $0.97 \pm 0.04$ & $0.99 \pm 0.00$ & $1.00 \pm 0.00$ & $0.17 \pm 0.00$ & $1.00 \pm 0.00$ & $0.86 \pm 0.06$ & $0.96 \pm 0.03$\\
\rowcolor[HTML]{EFEFEF} 10 & $0.22 \pm 0.06$ & $0.60 \pm 0.06$ & $0.81 \pm 0.05$ & $0.53 \pm 0.06$ & $0.67 \pm 0.06$ & $0.93 \pm 0.07$ & $0.89 \pm 0.05$ & $0.99 \pm 0.03$ & $0.08 \pm 0.00$ & $1.00 \pm 0.00$ & $0.77 \pm 0.06$ & $0.78 \pm 0.07$\\
15 & $0.21 \pm 0.05$ & $0.55 \pm 0.05$ & $0.83 \pm 0.04$ & $0.34 \pm 0.04$ & $0.72  \pm 0.05$ & $0.84 \pm 0.05$ & $0.95 \pm 0.02$ & $0.99 \pm 0.00$ & $0.16 \pm 0.00$ & $1.00 \pm 0.00$ & $0.77 \pm 0.06$ & $0.81 \pm 0.06$ \\
\rowcolor[HTML]{EFEFEF} 50 & $0.07 \pm 0.17$ & $0.20 \pm 0.35$ & $0.39 \pm 0.31$ & $0.41 \pm 0.32$ & $0.75 \pm 0.24$ & $0.77 \pm 0.23$ & $0.85 \pm 0.18$ & $-$ & $-$ & $-$ &  $0.64 \pm 0.05$ & $0.87 \pm 0.05$ \\
100 & $0.05 \pm 0.10$ & $0.17 \pm 0.31$ & $0.27 \pm 0.24$ & $0.58 \pm 0.29$ & $0.78 \pm 0.15$ & $0.76 \pm 0.24$ & $0.70 \pm 0.23$ & $0.94 \pm 0.04$ & $0.87 \pm 0.00$ & $1.00 \pm 0.00$ & $0.59 \pm 0.05$ & $0.80 \pm 0.06$\\
\rowcolor[HTML]{EFEFEF}200 & $0.07 \pm 0.15$ & $0.06 \pm 0.06$ & $0.27 \pm 0.19$ & $0.68 \pm 0.11$ & $0.75 \pm 0.11$ & $0.86 \pm 0.19$ & $0.64 \pm 0.20$ & $-$ & $-$ & $-$ & $0.67 \pm 0.03$ & $0.82 \pm 0.06$\\
300 & $0.03 \pm 0.06$ & $0.09 \pm 0.14$ & $0.34 \pm 0.22$ & $0.71 \pm 0.13$ & $0.74 \pm 0.12$ & $0.81 \pm 0.22$ & $0.60 \pm 0.09$ & $0.87 \pm 0.08$ & $0.95 \pm 0.00$ & $1.00 \pm 0.00$ & $0.65 \pm 0.04$ & $0.67 \pm 0.07$\\
\rowcolor[HTML]{EFEFEF}400 & $0.04 \pm 0.05$ & $0.11 \pm 0.13$ & $0.43 \pm 0.21$ & $0.73 \pm 0.12$ & $0.72 \pm 0.09$ & $0.86 \pm 0.19$ & $0.66 \pm 0.15$ & $-$ & $-$ & $-$ & $0.62 \pm 0.04$ & $0.67 \pm 0.07$\\
600 & $0.10 \pm 0.08$ & $0.12 \pm 0.10$ & $0.59 \pm 0.18$ & $0.75 \pm 0.12$ & $0.74 \pm 0.09$ & $0.87 \pm 0.16$ & $0.68 \pm 0.13$ & $-$ & $-$ & $-$ & $0.69 \pm 0.02$ & $0.60 \pm 0.06$\\
\rowcolor[HTML]{EFEFEF}800 & $0.10 \pm 0.07$ & $0.16 \pm 0.09$ & $0.73 \pm 0.13$ & $0.74 \pm 0.08$ & $0.77 \pm 0.11$ & $0.85 \pm 0.16$ & $0.67 \pm 0.10$ & $-$ & $-$ & $-$ & $0.72 \pm 0.01$ & $0.71 \pm 0.05$ \\
1000 & $0.16 \pm 0.07$ & $0.20 \pm 0.11$ & $-$ & $0.77 \pm 0.10$ & $0.78 \pm 0.10$ & $0.91 \pm 0.15$ & $0.70 \pm 0.08$ & $-$ & $-$ & $-$ & $0.75 \pm 0.01$ & $0.64 \pm 0.05$\\
\rowcolor[HTML]{EFEFEF}1500 & $0.25 \pm 0.07$ & $0.33 \pm 0.11$ & $-$ & $0.80 \pm 0.10$ & $0.80 \pm 0.09$ & $0.91 \pm 0.15$ & $0.78 \pm 0.09$ & $-$ & $-$ & $-$ & $0.79 \pm 0.02$ & $0.63 \pm 0.04$\\
2000 & $0.30 \pm 0.08$ & $0.52 \pm 0.10$ & $-$ & $0.81 \pm 0.09$ & $0.81 \pm 0.08$ & $0.96 \pm 0.10$ & $0.80 \pm 0.09$ & $-$ & $-$ & $-$ & $0.76 \pm 0.01$ & $0.78 \pm 0.04$\\
\rowcolor[HTML]{EFEFEF}3000 & $0.41 \pm 0.10$ & $0.62 \pm 0.08$ & $-$ & $0.87 \pm 0.09$ & $-$ & $-$ & $-$ & $-$ & $-$ & $-$ & $0.79 \pm 0.02$ & $0.78 \pm 0.03$\\
4000 & $0.55 \pm 0.08$ & $0.70 \pm 0.03$ & $-$ & $0.93 \pm 0.08$ & $-$ & $-$ & $-$ & $-$ & $-$ & $-$ &  $0.79 \pm 0.02$ & $0.81 \pm 0.03$\\ 
\rowcolor[HTML]{EFEFEF}5000 & $0.65 \pm 0.08$ & $0.76 \pm 0.03$ & $-$ & $0.94 \pm 0.04$ & $-$ & $-$ & $-$ & $-$ & $-$ & $-$ & $0.81 \pm 0.01$ & $-$\\
\midrule
\multicolumn{11}{c}{FR} \\ \midrule
5 & $0.08 \pm 0.04$ & $0.74 \pm 0.08$ & $0.99 \pm 0.01$ & $0.81 \pm 0.07$ & $0.93 \pm 0.04$ & $0.94 \pm 0.05$ & $1.00 \pm 0.00$ & $1.00 \pm 0.00$ & $1.00 \pm 0.00$ & $1.00 \pm 0.00$ & $0.91 \pm 0.04$ & $0.96 \pm 0.04$\\
\rowcolor[HTML]{EFEFEF}10 & $0.12 \pm 0.06$ & $0.37 \pm 0.09$ &  $0.90 \pm 0.05$ & $0.51 \pm 0.09$ & $0.75 \pm 0.07$ & $0.90 \pm 0.04$ & $0.94 \pm 0.07$ & $0.99 \pm 0.00$ & $1.00 \pm 0.00$ & $1.00 \pm 0.00$ & $0.82 \pm 0.07$ & $0.86 \pm 0.07$\\
15 & $0.22 \pm 0.07$ & $0.33 \pm 0.08$ & $0.79 \pm 0.06$ & $0.44 \pm 0.09$ & $0.75 \pm 0.07$ & $0.77 \pm 0.05$ & $0.83 \pm 0.09$ & $0.99 \pm 0.00$ & $1.00 \pm 0.00$ & $1.00 \pm 0.00$ & $0.73 \pm 0.07$ & $0.87 \pm 0.05$\\
\rowcolor[HTML]{EFEFEF}50 & $0.06 \pm 0.08$ & $0.07 \pm 0.09$ & $0.38 \pm 0.25$ & $0.35 \pm 0.28$ & $0.78 \pm 0.23$ & $0.73 \pm 0.27$ & $0.73 \pm 0.29$ & $-$ & $-$ & $-$ & $0.74 \pm 0.05$ & $0.83 \pm 0.08$\\
100 & $0.02 \pm 0.02$ & $0.04 \pm 0.06$ & $0.38 \pm 0.21$ & $0.46 \pm 0.22$ & $0.73 \pm 0.19$ & $0.60 \pm 0.21$ & $0.72 \pm 0.21$ & $0.95 \pm 0.00$ & $1.00 \pm 0.00$ & $1.00 \pm 0.00$ & $0.60 \pm 0.06$ & $0.73 \pm 0.08$\\
\rowcolor[HTML]{EFEFEF}200 & $0.03 \pm 0.02$ & $0.09 \pm 0.19$ & $0.30 \pm 0.12$ & $0.67 \pm 0.13$ & $0.71 \pm 0.13$ & $0.81 \pm 0.21$ & $0.68 \pm 0.19$ & $-$ & $-$ & $-$ & $0.82 \pm 0.03$ & $0.63 \pm 0.08$\\
300 & $0.06 \pm 0.06$ & $0.07 \pm 0.09$ & $0.43 \pm 0.21$ & $0.69 \pm 0.10$ & $0.70 \pm 0.07$ & $0.80 \pm 0.19$ & $0.74 \pm 0.20$ & $1.00 \pm 0.00$ & $1.00 \pm 0.00$ & $1.00 \pm 0.00$ & $0.78 \pm 0.04$ & $0.87 \pm 0.05$\\
\rowcolor[HTML]{EFEFEF}400 & $0.06 \pm 0.05$ & $0.12 \pm 0.22$ & $0.51 \pm 0.22$ & $0.68 \pm 0.07$ & $0.68 \pm 0.07$ & $0.88 \pm 0.18$ & $0.70 \pm 0.12$ & $-$ & $-$ & $-$ & $0.79 \pm 0.04$ & $0.63 \pm 0.09$\\
600 & $0.13 \pm 0.06$ & $0.10 \pm 0.08$ & $0.64 \pm 0.14$ & $0.78 \pm 0.12$ & $0.77 \pm 0.10$ & $0.84 \pm 0.17$ & $0.68 \pm 0.06$ & $-$ & $-$ & $-$ &  $0.77 \pm 0.02$ & $0.74 \pm 0.07$\\
\rowcolor[HTML]{EFEFEF}800 & $0.18 \pm 0.05$ & $0.14 \pm 0.08$ & $0.75 \pm 0.14$ & $0.77 \pm 0.11$ & $0.75 \pm 0.07$ & $0.88 \pm 0.16$ & $0.73 \pm 0.15$ & $-$ & $-$ & $-$ & $0.81 \pm 0.02$ & $0.61 \pm 0.07$\\
1000 & $0.21 \pm 0.07$ & $0.20 \pm 0.08$ & $-$ & $0.78 \pm 0.13$ & $0.77 \pm 0.09$ & $0.88 \pm 0.15$ & $0.76 \pm 0.11$ & $-$ & $-$ & $-$ & $0.81 \pm 0.02$ & $0.74 \pm 0.07$\\
\rowcolor[HTML]{EFEFEF}1500 & $0.33 \pm 0.07$ & $0.38 \pm 0.07$ & $-$ & $0.83 \pm 0.12$ & $0.80 \pm 0.07$ & $0.93 \pm 0.12$ & $0.78 \pm 0.09$ & $-$ & $-$ & $-$ &  $0.82 \pm 0.02$ & $0.69 \pm 0.05$ \\
2000 & $0.37 \pm 0.07$ & $0.55 \pm 0.06$ & $-$ & $0.82 \pm 0.10$ & $0.86 \pm 0.09$ & $0.90 \pm 0.13$ & $0.79 \pm 0.07$ & $-$ & $-$ & $-$ & $0.78 \pm 0.04$ & $0.69 \pm 0.04$\\
\rowcolor[HTML]{EFEFEF}3000 & $0.48 \pm 0.07$ & $0.67 \pm 0.05$ & $-$ & $0.84 \pm 0.10$ & $-$ & $-$ & $-$ & $-$ & $-$ & $-$ & $0.84 \pm 0.02$ & $0.77 \pm 0.03$ \\
4000 & $0.60 \pm 0.08$ & $0.71 \pm 0.03$ & $-$ & $0.92 \pm 0.07$ & $-$ & $-$ & $-$ & $-$ & $-$ & $-$ & $0.86 \pm 0.02$ & $0.82 \pm 0.02$\\
\rowcolor[HTML]{EFEFEF}5000 & $0.67 \pm 0.07$ & $0.79 \pm 0.09$ & $-$ & $0.88 \pm 0.02$ & $-$ & $-$ & $-$ & $-$ & $-$ & $-$ & $0.87 \pm 0.02$ & $-$\\
\midrule
\multicolumn{11}{c}{ZH} \\\midrule
5 & $0.53 \pm 0.04$ & $0.74 \pm 0.05$ & $0.98 \pm 0.01$ & $0.98 \pm 0.01$ & $0.94 \pm 0.02$ & $1.00 \pm 0.00$ & $1.00 \pm 0.07$ & $1.00 \pm 0.00$ & $1.00 \pm 0.00$ & $1.00 \pm 0.00$  & $0.97 \pm 0.02$ & $0.95 \pm 0.02$\\
\rowcolor[HTML]{EFEFEF}10 & $0.54 \pm 0.02$ & $0.64 \pm 0.03$  & $0.97 \pm 0.01$ & $0.92 \pm 0.04$ & $0.94 \pm 0.01$ & $0.98 \pm 0.00$ & $1.00 \pm 0.06$ & $1.00 \pm 0.00$ & $1.00 \pm 0.00$ & $1.00 \pm 0.00$ & $0.93 \pm 0.03$ & $1.00 \pm 0.00$\\
15 & $0.53 \pm 0.02$ & $0.67 \pm 0.03$ & $0.99 \pm 0.00$ & $0.93 \pm 0.03$ & $0.93 \pm 0.02$ & $0.99 \pm 0.00$ & $0.95 \pm 0.10$ & $1.00 \pm 0.00$ & $1.00 \pm 0.00$ & $1.00 \pm 0.00$ & $0.91 \pm 0.05$ & $0.99 \pm 0.00$\\
\rowcolor[HTML]{EFEFEF}50 & $0.58 \pm 0.10$ & $0.71 \pm 0.13$ & $0.96 \pm 0.14$ & $0.94 \pm 0.21$ & $0.94 \pm 0.14$ & $0.99 \pm 0.01$ & $0.98 \pm 0.02$ & $-$ & $-$ & $-$ & $0.97 \pm 0.01$ & $0.95 \pm 0.02$\\
100 & $0.57 \pm 0.06$ & $0.71 \pm 0.12$ & $0.97 \pm 0.07$ & $0.94 \pm 0.09$ & $0.95 \pm 0.04$ & $0.97 \pm 0.03$ & $0.97 \pm 0.05$ & $1.00 \pm 0.00$ & $1.00 \pm 0.00$ & $1.00 \pm 0.00$ &   $0.98 \pm 0.01$ & $0.96 \pm 0.02$\\
\rowcolor[HTML]{EFEFEF}200 & $0.62 \pm 0.09$ & $0.67 \pm 0.08$ & $0.98 \pm 0.05$ & $0.95 \pm 0.05$ & $0.96 \pm 0.03$ & $0.95 \pm 0.05$ & $0.97 \pm 0.04$ & $-$ & $-$ & $-$ & $0.97 \pm 0.01$ & $0.93 \pm 0.02$\\
300 & $0.66 \pm 0.07$ & $0.73 \pm 0.10$ & $0.99 \pm 0.02$ & $0.96 \pm 0.03$ & $0.95 \pm 0.05$ & $0.95 \pm 0.04$ & $0.98 \pm 0.02$ & $1.00 \pm 0.00$ & $1.00 \pm 0.00$ & $1.00 \pm 0.00$ & $0.97 \pm 0.00$ & $0.94 \pm 0.01$\\
\rowcolor[HTML]{EFEFEF}400 & $0.70 \pm 0.06$ & $0.76 \pm 0.08$ & $0.92 \pm 0.05$ & $0.96 \pm 0.03$ & $0.96 \pm 0.05$ & $0.95 \pm 0.04$ & $0.98 \pm 0.02$ & $-$ & $-$ & $-$ & $0.97 \pm 0.01$ & $0.96 \pm 0.01$\\
600 & $0.80 \pm 0.11$ & $0.85 \pm 0.06$ & $-$ & $0.95 \pm 0.03$ & $0.98 \pm 0.02$ & $0.95 \pm 0.03$ & $0.97 \pm 0.03$ & $-$ & $-$ & $-$ & $0.97 \pm 0.01$ & $0.95 \pm 0.02$\\
\rowcolor[HTML]{EFEFEF}800 & $0.85 \pm 0.08$ & $0.91 \pm 0.07$ & $-$ & $0.96 \pm 0.03$ & $0.97 \pm 0.02$ & $0.93 \pm 0.07$ & $0.96 \pm 0.03$ & $-$ & $-$ & $-$  &  $0.97 \pm 0.01$ & $0.96 \pm 0.01$\\
1000 & $0.91 \pm 0.09$ & $0.95 \pm 0.06$ & $-$ & $0.96 \pm 0.03$ & $0.96 \pm 0.02$ & $0.96 \pm 0.03$ & $0.98 \pm 0.03$ & $-$ & $-$ & $-$ & $0.98 \pm 0.01$ & $0.96 \pm 0.01$\\
\rowcolor[HTML]{EFEFEF}1500 & $0.91 \pm 0.10$ & $0.95 \pm 0.06$ & $-$ & $0.96 \pm 0.04$ & $0.97 \pm 0.02$ & $0.96 \pm 0.03$ & $0.95 \pm 0.05$ & $-$ & $-$ & $-$ & $0.98 \pm 0.01$ & $0.95 \pm 0.01$\\
2000 & $0.91 \pm 0.10$ & $0.96 \pm 0.05$ & $-$ & $0.97 \pm 0.04$ & $0.98 \pm 0.02$ & $0.97 \pm 0.03$ & $0.97 \pm 0.03$ & $-$ & $-$ & $-$ & $0.98 \pm 0.01$ & $0.94 \pm 0.02$\\
\rowcolor[HTML]{EFEFEF}3000 & $0.92 \pm 0.07$ & $0.96 \pm 0.05$ & $-$ & $0.98 \pm 0.03$ & $-$ & $-$ & $-$ & $-$ & $-$ & $-$ & $0.99 \pm 0.00$ & $0.96 \pm 0.01$\\
4000 & $0.92 \pm 0.06$ & $0.98 \pm 0.02$ & $-$ & $0.97 \pm 0.02$ & $-$ & $-$ & $-$ & $-$ & $-$ & $-$ & $0.98 \pm 0.01$ & $0.96 \pm 0.01$\\
\rowcolor[HTML]{EFEFEF}5000 & $0.94 \pm 0.04$ & $0.99 \pm 0.02$ & $-$ & $0.98 \pm 0.01$ & $-$ & $-$ & $-$ & $-$ & $-$ & $-$ & $0.97 \pm 0.01$ & $-$\\
\bottomrule
\end{tabular}%
} 
\caption{Normalized Edit Distance (NED) scores for multilingual text rendering across various text lengths. Each model is prompted to generate an image embedding ground truth text sampled from Wikipedia~\cite{wikidump} in English, French, or Chinese. OCR is applied to the generated images, and NED is computed between the OCR output and the ground-truth text. Lower scores indicate higher fidelity in character-level text rendering.}
\label{tab:ned}
\end{table*}

\newpage

\section{Appendix: CER and WER Figure}
\begin{figure*}[t]
    \centering
    \includegraphics[width=1\linewidth]{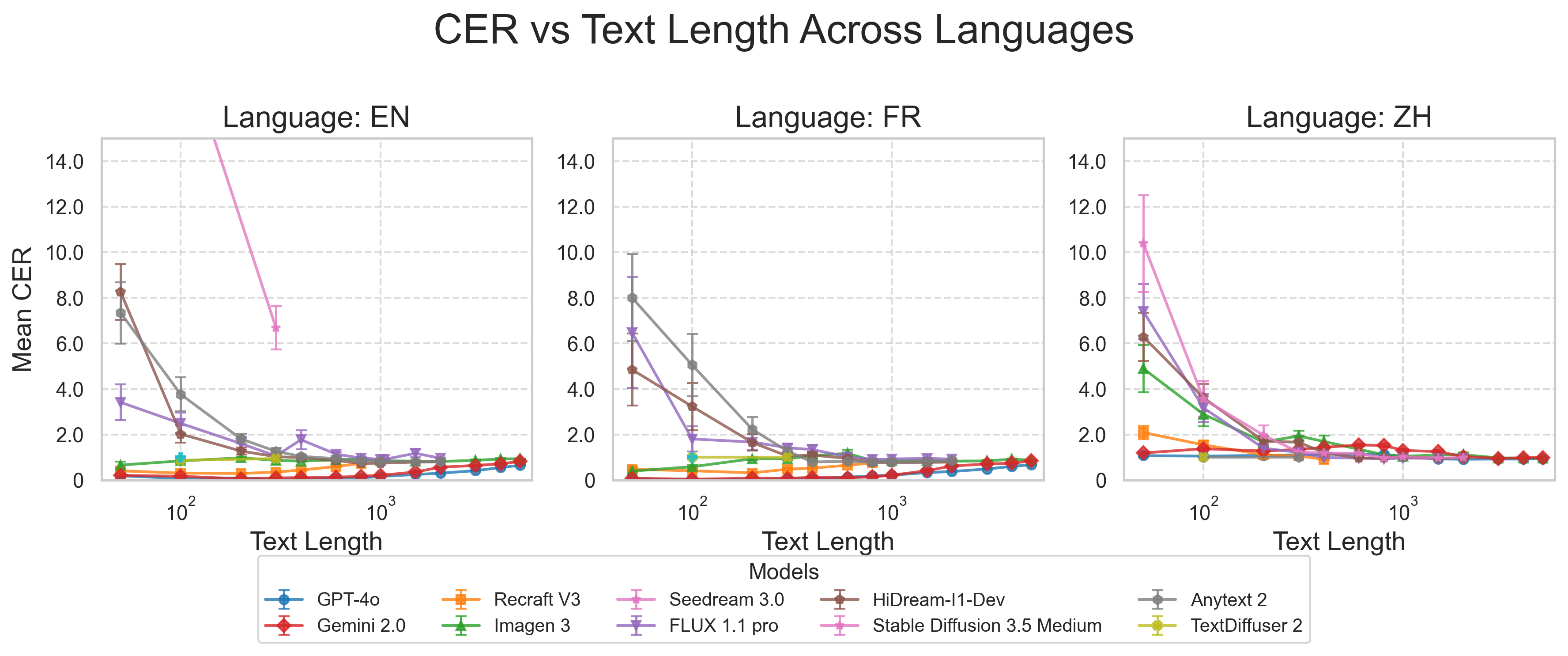}
    
\caption{
Character Error Rate (CER) vs. Text Length across Languages.
    We evaluate ten state-of-the-art text-to-image generation models on multilingual text rendering using English (EN), French (FR), and Chinese (ZH) excerpts sampled from Wikipedia, with input lengths ranging from 5 to 5000 characters. Each model is prompted with identical semantic content across varying lengths, and OCR is applied to the generated images to compute Character Error Rate (CER). Higher-performing models such as GPT-4o, Gemini 2.0, and Imagen 3 are evaluated up to 5000 characters, while Stable Diffusion 3.5, AnyText2, and TextDiffuser2 are evaluated up to 300 characters, and the remaining models up to 2000. Lower CER scores indicate better text fidelity and layout consistency.
}
\label{fig:cer}
\end{figure*}

\begin{figure*}[t]
    \centering
    \includegraphics[width=1\linewidth]{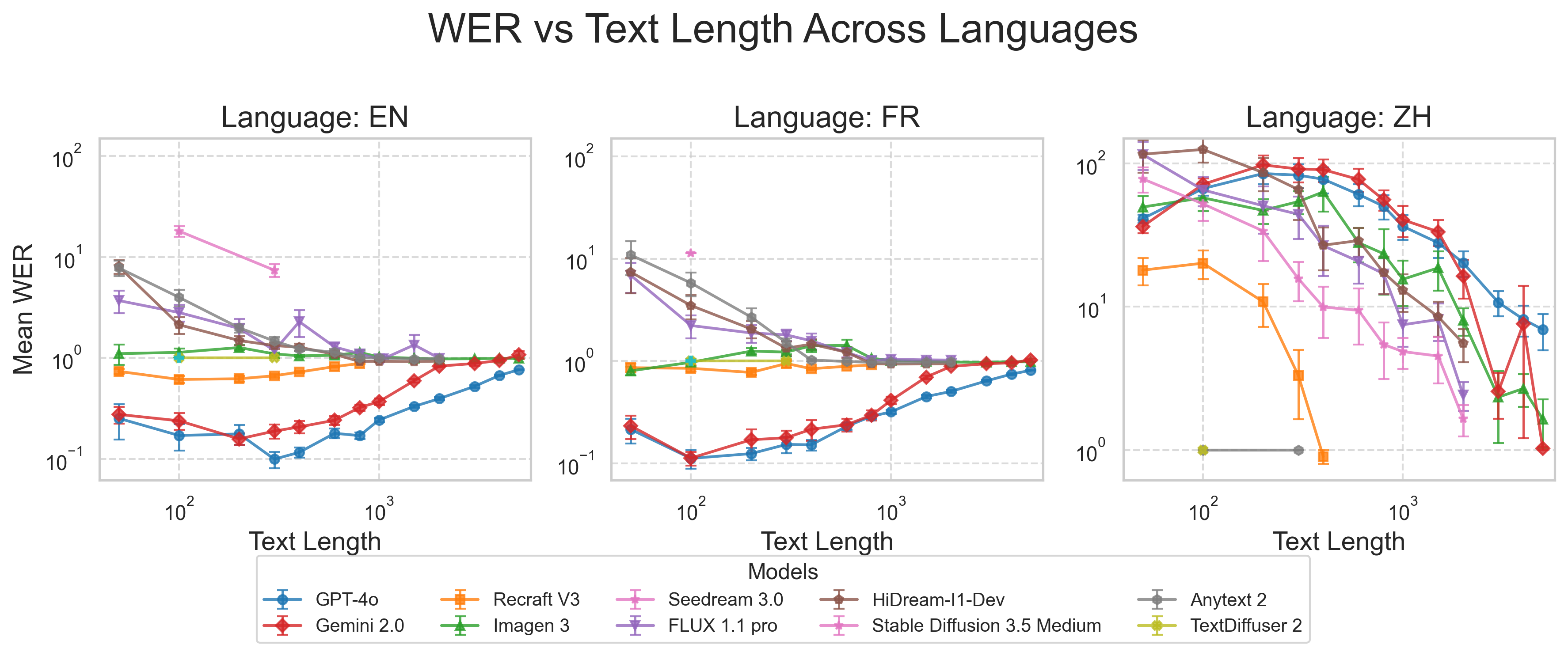}

\caption{
Word Error Rate (WER) vs. Text Length across Languages.
    We evaluate ten state-of-the-art text-to-image generation models on multilingual text rendering using English (EN), French (FR), and Chinese (ZH) excerpts sampled from Wikipedia, with input lengths ranging from 5 to 5000 characters. Each model is prompted with identical semantic content across varying lengths, and OCR is applied to the generated images to compute Word Error Rate (WER). Higher-performing models such as GPT-4o, Gemini 2.0, and Imagen 3 are evaluated up to 5000 characters, while Stable Diffusion 3.5, AnyText2, and TextDiffuser2 are evaluated up to 300 characters, and the remaining models up to 2000. Lower WER scores indicate better text fidelity and layout consistency.
}
    \label{fig:wer}
\end{figure*}

Please check Figure \ref{fig:cer} and Figure \ref{fig:wer} for Character Error Rate (CER) and Word Error Rate (WER).

 \label{sec:appendix2}

\section{Use of AI Tools in Manuscript Preparation}
\label{sec:appendix_ai}
In the preparation of this manuscript, we utilized a large language model as an assistive tool. The LLM's role was confined to improving the grammatical structure and clarity of our writing. Furthermore, it was used to assist in debugging code snippets and generating routine documentation such as docstrings. The core research ideas, experimental design, analysis, and conclusions were conceived and executed entirely by the authors. All LLM-generated outputs were critically reviewed and edited by the authors, who take full responsibility for the final content of this paper.
\end{document}